%% file: lexalign-cogsci2024.tex
\documentclass[10pt,letterpaper]{article}
\usepackage[table]{xcolor}%
\usepackage{colortbl}
\usepackage{tabularx}
\usepackage{multirow}
\usepackage{multicol}
\usepackage{cogsci}
\usepackage{array}
\cogscifinalcopy %
\usepackage{subcaption}
\usepackage{graphicx}
\usepackage{booktabs}
\usepackage{pslatex} 
\usepackage{color}
\usepackage{float} %

\usepackage{amsmath}
\usepackage{xurl}
\usepackage{hyperref}
\usepackage[backend=biber, maxbibnames=100, style=apa]{biblatex}

\usepackage[style=apa, backend=biber]{biblatex}
\definecolor{mycolor}{rgb}{0.0,0.0,1} %
\definecolor{rfcolor}{rgb}{0.0,0.5,0.0}
\definecolor{wpcolor}{rgb}{0.9,0.0,0.0} 
\definecolor{newcolor}{rgb}{0.0,0.0,0.9} 
\definecolor{mrcolor}{rgb}{0.5,0.5,0.0}

\title{Analysing Cross-Speaker Convergence in Face-to-Face Dialogue through\\ the Lens of Automatically Detected Shared Linguistic Constructions} 

\author{{\large \bf Esam Ghaleb (e.ghaleb@uva.nl)} \\
 Institute for Logic, Language and Computation, University of Amsterdam, The Netherlands
  \AND {\large \bf Marlou Rasenberg (marlou.rasenberg@meertens.knaw.nl)} \\
  Meertens Institute, Royal Netherlands Academy of Arts and Sciences, The Netherlands
  \AND{\large \bf Wim Pouw (wim.pouw@ru.nl)} \& {\large \bf Ivan Toni (ivan.toni@ru.nl)}\\
    Donders Institute for Brain, Cognition, and Behaviour, Radboud University, The Netherlands
  \AND {\large \bf Judith Holler (judith.holler@mpi.nl)} \&
  {\large \bf Asl\i~Özy\"{u}rek (asli.ozyurek@mpi.nl)} \\
    Donders Institute for Brain, Cognition, and Behaviour, Radboud University \& MPI for Psycholinguistics, The Netherlands
  \AND {\large \bf Raquel Fernández (raquel.fernandez@uva.nl)} \\
  Institute for Logic, Language and Computation, University of Amsterdam, The Netherlands}

\begin{document}

\maketitle

\begin{abstract}
Conversation requires a substantial amount of coordination between dialogue participants, from managing turn taking to negotiating mutual understanding. Part of this coordination effort surfaces as the reuse of linguistic behaviour across speakers, a process often referred to as \textit{alignment}. While the presence of linguistic alignment is well documented in the literature, several questions remain open, including the extent to which patterns of reuse across speakers have an impact on the emergence of labelling conventions for novel referents. In this study, we put forward a methodology for automatically detecting \textit{shared lemmatised constructions}---expressions with a common lexical core used by both speakers within a dialogue---and apply it to a referential communication corpus where participants aim to identify novel objects for which no established labels exist. Our analyses uncover the usage patterns of shared constructions in interaction and reveal that features such as their frequency and the amount of different constructions used for a referent are associated with the degree of object labelling convergence the participants exhibit after social interaction.
More generally, the present study shows that automatically detected shared constructions offer a useful level of analysis to investigate the dynamics of reference negotiation in dialogue.

\end{abstract}

\input{sections/1-introduction}

\input{sections/2-methods}

\input{sections/3-analysis1}

\input{sections/4-analysis2}

\input{sections/5-analysis3}

\input{sections/6-discussion}
\input{sections/7-conclusion}

\section{Acknowledgments}
This research was funded by the Dutch Research Council (NWO) under a Gravitation grant (024.001.006) awarded to the Language in Interaction consortium. Raquel Fern\'andez and Judith Holler are supported by the European Research Council (ERC grant agreement numbers 819455 and 773079, respectively).

\setlength{\bibhang}{.125in}

\printbibliography

\end{document}

%% file: sections/1-introduction.tex
\section{Introduction}

Speakers use language flexibly in conversation in order to negotiate mutual understanding and achieve joint goals \autocite{clark1996using}.  
Repeated referential games have provided a framework to study such coordination processes \autocite{krauss1966concurrent,clark1986referring,fay2018create,motamedi2019evolving}.
These games typically involve several rounds where one participant (the director) describes a novel object to another one (the matcher), who needs to identify it among several candidates. 
In these referential director-matcher games, pairs are faced with two major challenges. First, the director and matcher need to work together so that the matcher can find the target object, which is a collaborative undertaking \autocite{clark1986referring}. To this end, the director often recruits detailed %
descriptions of the object (e.g., ``it looks like a triangle shape on top'' rather than shorter descriptions like ``pyramid''). Yet it is only after the matcher has provided positive evidence of understanding (e.g., ``ok I got it'') or demonstrated understanding (``oh yeah like a pyramid'') that the referential expression has been grounded \autocite{clark1991grounding}. After this grounding process, the pairs over multiple referrals start to devise more economical and efficient references to the objects, either due to natural proclivities of efficient communication and/or due to pressures to complete the game in a fast way. The challenge here is to construct an economical and portable reference that somehow allows the pair to easily identify the referent, overcoming limitations of memory, ambiguity, semantic transparency, and likely other communication-typical bottlenecks. This second challenge is often met by using references that derive from the initially grounded referential expressions (e.g., ``pyramid''), thus leading to alignment \autocite{brennan1996conceptual,pickering2004toward}. 

The detailed process and consequences of the pair settling or choosing a certain portable referential expression to be used throughout future interactions are not well-studied and certainly not often studied in a way that is computationally reproducible and automatable. This is an important gap in the literature, as it is precisely this settling or sifting process, where references become entrained through reuse in interaction, that is likely an important aspect of how languages evolve (to this day) and how conventionalization unfolds. Being able to automatically track the re-use of referential expressions in natural conversation promises to provide a novel lens through which to advance our understanding of these core-linguistic dynamics of reuse.

Methodologically, several 
issues arise for tracking joint reuse. All types of words are reused all the time in conversation (for the same language is spoken), and it is not immediately clear how to %
decide what shared vs.~non-shared references are.
This issue needs to be addressed by finding the right level of abstraction and constructing the right selection criteria for the discovery of relevant reused material within %
`messy' conversation.
There have been several proposals for automatically tracking different types of linguistic reuse, including syntactic and lexical levels of representation \autocite{nenkova2008high,fernandez2014quantifying,healey2014divergence,duran2019align,dideriksen2023quantifying}.
Below, we present what we believe is a computationally effective %
way to automate the process of discovering \textit{shared constructions} where participants align in form and meaning. 
With these methodology %
in hand, we address the following key questions. Broadly, we aim to understand when and how often shared constructions arise and whether such constructions appear to mediate 
the outcomes of the interaction (i.e., the establishment of naming conventions as measured after social interaction). We conjecture that share constructions provide a useful level of analysis to investigate the dynamics of reference negotiation in interaction, and hypothesize that their properties and patterns of use are directly associated with the level of convergence participants exhibit after having interacted.

To answer these questions, we analyse a corpus of $66$ dialogues where participants with director-matcher roles engage in a repeated referential task with novel objects, and also perform an individual naming task for each novel object before and after the interaction. Using this data and our method to automatically detect shared constructions, we conduct three analyses. Analysis 1 examines how often %
shared constructions arise and the degree of variation within the shared constructions that emerge for a referent. We put our method to the test by devising a strong baseline using pseudo-pair conversations. These pseudo-pairs help us differentiate between shared constructions that might simply arise because the objects invite certain ways of referring to them versus shared constructions that arise because of the pair-unique conversational history that is built over time.
Analysis 2 studies the relationship between shared constructions and individual speaker labels for objects before and after the interaction.
Finally, Analysis 3 investigates whether the features of shared constructions can explain the speakers' convergence when labelling objects following their interaction. 

Together, these analyses shed new light on the dynamics of cross-speaker linguistic convergence, using an automated method suitable for quantifying convergence in large dialogue corpora.

%% file: sections/2-methods.tex
\section{Methods}

 \subsection{Dataset}
Our study utilises transcribed conversations from 66 pairs of Dutch-speaking participants. Concretely, the data we use comes from two multimodal datasets: 19 dyads are part of the dataset introduced by \textcite{rasenberg2022primacy}, and 47 dyads are part of the CABB dataset \autocite{eijk2022cabb}, but participants in both datasets completed the same referential director-matcher task. Overall, the combined corpus includes 27 all-female, 10 all-male, and 29 mixed-gender dyads, with participant ages ranging from 18 to 33 years old (22.6 average). Participants were not acquainted with each other beforehand.

\begin{figure}[ht!]
    \centering
    \includegraphics[width=0.99\linewidth]{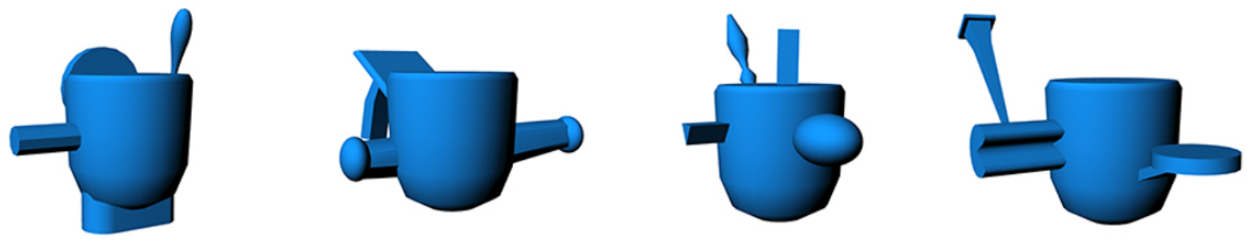}
    \caption{Four example fribbles used as stimuli in the task.}
    \label{fig:fribble01}
\end{figure}

In this dataset, participants alternate between director and matcher roles to jointly identify images of 16 novel 3D objects called ``fribbles'' (see Figure \ref{fig:fribble01}), first proposed by \textcite{barry2014meet}. They do so for a total of six rounds. In each round, the director describes a fribble for the matcher to identify among the 16 candidates. The 16 fribbles are displayed on a screen (one screen per participant, visually inaccessible to the respective other) and are distributed randomly over 16 positions in each round. The performance of the pairs in resolving references is near the ceiling (the mean accuracy is 99.4\%). On average, the interactive task takes 25 minutes. 

Before and after the collaborative, interactive task, speakers participate in an individual naming task, where they are asked to name each fribble with 1 to 3 words that would allow their partner to identify it amongst the other fribbles. These names are not produced within dialogue interaction and hence reflect the lexical labels that the individual participants associate with the objects before and after the communicative task.

\begin{figure*}[ht!]
    \centering    \includegraphics[width=0.99\linewidth]{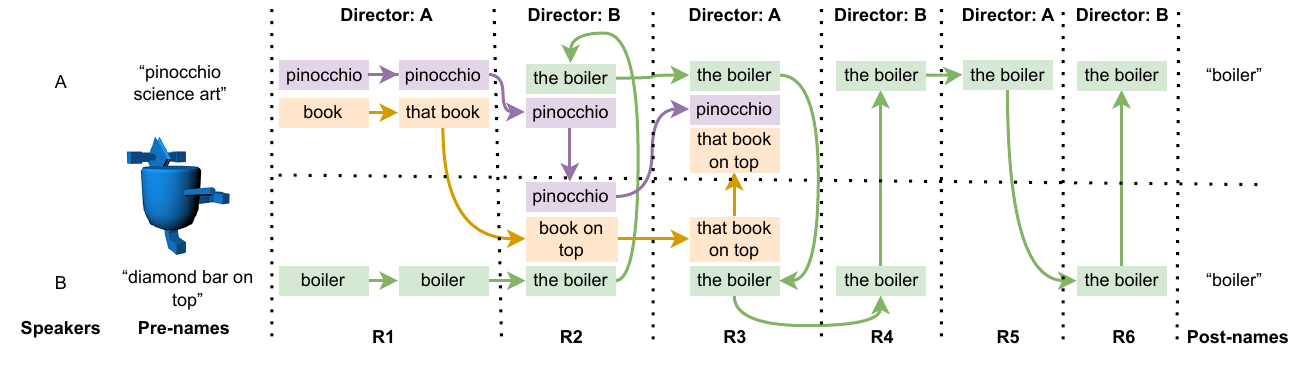}
    \caption{This figure shows the pre- and post-interaction names and the shared construction types for a fribble by a pair. Before the interaction, speakers A and B refer to the fribble as ``pinocchio science art'' and ``diamond bar on top,'' respectively. The figure shows the shared constructions that emerge, with the arrows indicating the order in which speakers repeat these constructions. For instance, in the first round, speaker A, acting as the director, refers to the fribble as ``pinocchio'' twice, and speaker B repeats this construction in the second round. 
    This dyad uses three shared construction types for this fribble (indicated by the colours purple, orange, and green). The 
    types ``book'' and  ``pinocchio'' are dropped after Round 3, while the type ``boiler'' 
    is used in all rounds (a total of 11 times) as well as in the post-interaction names by both speakers. }
    \label{fig:shared_expression_types}
\end{figure*}

\subsection{Extracting Shared Lemmatised Constructions}
We are interested in capturing the reuse of linguistic behaviour across dialogue partners when they refer to the same object; hence in instances including alignment of both form and meaning \autocite{rasenberg2020alignment}. Regarding form, we focus on cross-speaker matching of lemmatised speech, i.e., speech where content words (nouns, adjectives, verbs, and adverbs) are reduced to their base form. For example, the Dutch word ``ball'' (singular noun), ``ballen'' (plural noun) and ``balletje'' (noun with diminutive suffix) can all be reduced to the lemma ``ball'' (ball). This captures a level of representation that is more abstract than lexical alignment with exact word matching while still being semantic (in contrast to syntactic alignment). We believe that cross-speaker matches at the level of lemmatised content word constructions that refer to the same object can be good candidates for capturing ‘conceptual pacts’ \autocite{brennan1996conceptual}, i.e., dialogue-based agreements about how to refer to an object.

Inspired by the approach of \textcite{sinclair-fernandez-2021-construction}, who (unlike us) focus on purely form-based lexical and syntactic alignment in the context of language acquisition, we propose a method to automatically identify the alignment of lemmatised constructions with a common referent. We adapt their computational method to extract matching sequences from dialogue transcripts to our data and purposes as follows.

Using the Python library spaCy \autocite{spacy2}, we first remove disfluencies, tag each word with its part of speech category, and lemmatise it to its base form. Next, for each dyad and each fribble, we combine all trials of the
dialogue (across rounds) where the participants are trying to identify that fribble. Using the sequential pattern-matching algorithm proposed by \textcite{dubuisson2021towards}, we extract all the sequences of lemmas (including single lemmas) that both dialogue participants have used across these sections of the dialogue. Finally, we filter out the sequences consisting exclusively of function words as well as those that are used for multiple fribbles, because they tend to correspond to generic phrases used across the board (e.g., ``the main part'' or ``head'').\footnote{Our code and data are available on GitHub at \href{https://github.com/EsamGhaleb/SharedLinguisticConstructions}{https://github.com/EsamGhaleb/SharedLinguisticConstructions}}.

This procedure results in a set of \textit{shared lemmatised constructions} (or \textit{shared constructions} for short) per dyad and fribble. These constructions can be further grouped into \textbf{\emph{shared construction types}} based on their common content lemmas. For instance, in the example shown in Figure~\ref{fig:shared_expression_types}, we group the shared constructions ``dat boek bovenop (that book on top)'', ``boek bovenop (book on top)'', ``dat boek (that book)'', and ``boek (book)'' uttered by both participants of a dyad to refer to a given fribble into the shared construction type ``book''.
As a result, our approach detects shared constructions with a common lexical core per referent throughout the entire dialogue, in contrast to other automated methods such as the ALIGN package \autocite{duran2019align}, which captures linguistic alignment as turn-by-turn overlap.

\subsection{Pseudo-Pairs}
To establish a baseline for our study, we create a dataset of 66 control dialogues constructed by combining speech by two participants from two different dialogues within the corpus, respecting the same director-matcher roles as in the original dialogues and using speech where the two). 
combined speakers refer to the same object.
However, unlike actual participant pairs, these pseudo-pairs did not directly interact in the referential communication task. Pseudo-pairs allow us to control for ``shared'' constructions that may arise by chance (e.g., because all participants refer to the same set of objects) rather than as a result of interaction.

\subsection{Computation of Lexical Cosine Similarity}
To measure the degree of form-based similarity between the names that the speakers produce in the individual naming task and between these names and shared lemmatised constructions, we use the same method as \textcite{rasenberg2022primacy}—lexical cosine similarity. To measure the similarity between two expressions, we compute the cosine similarity between the corresponding expression vectors with binary values (1/0), indicating whether a given lemma is present. Effectively, this measure captures lemma overlap while controlling for expression length, resulting in a similarity score ranging from 0 (no similarity) to 1 (full overlap). For example, the cosine similarity between ``pinocchio nose above'' and  ``window pinocchio nose'' is 0.67. Naming cosine similarity will become relevant in Analyses 2 \& 3.

%% file: sections/3-analysis1.tex
\section{Analysis 1: Presence of Shared Constructions and their Patterns of Use over Dialogue Rounds}
We start by quantifying the degree of linguistic alignment and how this evolves over a dialogue. Using the method described above, we automatically extract all the shared constructions and shared construction types for all dyads and all fribbles over all the six rounds of each dialogue. We do this for the 66 dialogues that make up the corpus as well as for the pseudo-pairs. This analysis provides information about the dynamics of shared constructions as well as a confirmation that shared constructions emerge in conversation (rather than simply being invited by the object's characteristics).

\begin{figure}[!ht]
    \centering
    \includegraphics[width=0.99\linewidth]{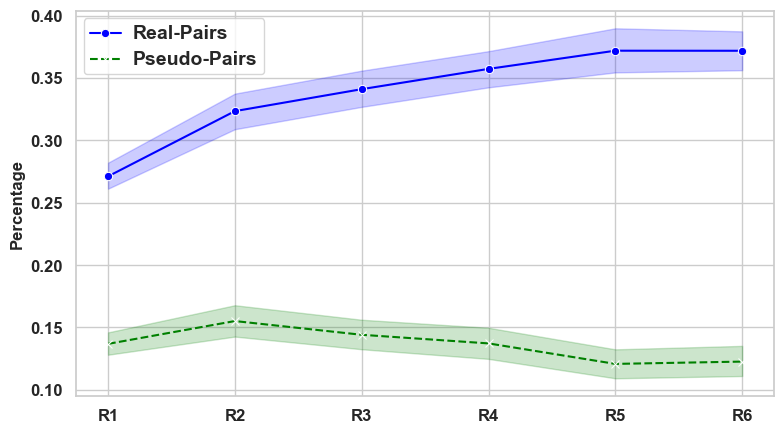}
    \caption{Percentage of utterances containing shared constructions over the rounds of interaction.}
    \label{fig:percentage_of_utterances_containing_shared_exp}
\end{figure}

We find that 92\% of participant pairs (61 out of 66) use at least one shared construction for each fribble. 
Furthermore, an average of 34\% of all utterances per dialogue include shared constructions. Figure \ref{fig:percentage_of_utterances_containing_shared_exp} shows that this rate increases as the interaction progresses: from 27\% in the first round to 37\% of utterances containing shared constructions in the final round (Spearman’s $\rho = 0.36, p \ll 0.001$). In contrast, only 14\% of utterances contain shared constructions in the pseudo-dialogues, with no increase over rounds. This indicates that alignment is largely the result of interaction, rather than solely the consequence of all dyads referring to the same objects.

Participant pairs tend to use several shared constructions per fribble. We find that, on average, dyads use 4 shared construction types per fribble. Different ways of referring to the same object may arise because participants refer to parts of a fribble—for example, as illustrated in Figure \ref{fig:shared_expression_types}, they may refer to the top part of the displayed fribble as ``book''---or they may arise because participants entertain different conceptualisations, e.g., using ``pinocchio'' vs. ``boiler'' to holistically refer to the object as a whole.
Either way, participants tend to converge on a smaller number of shared construction types towards the end of the dialogue (see, e.g., Figure \ref{fig:shared_expression_types}, where only ``boiler'' is used for this fribble in the last three rounds of the task). We find that the average number of shared construction types per fribble in the first round is 4.25, significantly higher than the average of 1.86 present in the last round ($t=16.45, p \ll 0.001$).

Overall, from this initial analysis we conclude that (1) our method to automatically identify shared constructions captures interaction-driven linguistic alignment, as shown by the contrast in the rate of shared constructions in real-pairs vs.~pseudo-pairs; (2) in line with findings in earlier work, linguistic alignment is pervasive in the communicative interactive task, with shared constructions 
emerging for almost all fribbles and being present in 34\% of utterances; (3) participants initially use several shared constructions types per referent
and over time some of these constructions tend to be dropped, as common ground is built up. %

%% file: sections/4-analysis2.tex
\section{Analysis 2: Individual Speaker Names versus Interactive Shared Constructions}
In our second analysis, we investigate the relation between cross-speaker alignment in the interactive task, as captured by shared constructions, and the names given to the fribbles by each participant in the individual naming task before and after the interaction. This analysis provides key information about what pairs bring to the table before the interaction and %
whether shared constructions established in interaction %
are linked to referential labels used after the dialogue. 

We start by examining whether the way in which participants individually name the fribbles is altered after the interactive task. For this, we calculate cosine similarity between the pre- and post-interaction names given by a participant for each fribble. If a speaker used the same name before and after the dialogue, we would obtain a cosine similarity of 1. Instead, we find that the average cosine similarity is 0.27 (std = 0.24), indicating that participants tend to name fribbles differently after the communicative task.

\begin{figure}[ht!]
    \centering
    \includegraphics[width=0.99\linewidth]{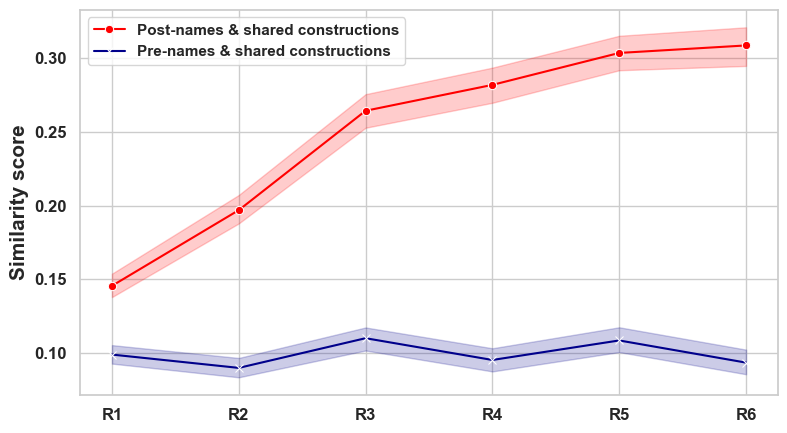}
    \caption{Lexical cosine similarity between a speaker's pre- and post-interaction names and the shared constructions used for a fribble, over the six dialogue rounds.}
    \label{fig:exp_vs_name_vs_utterances}
\end{figure}

To investigate to what extent this difference is mediated by interaction-based linguistic alignment, we compute the cosine similarity of the dyads' shared construction types for a fribble with the pre- and post-interaction names of each participant. 
On average, $41.3\%\pm{0.09}$ of pre-interaction names and $61.5\%\pm{0.10}$ of post-interaction names per participant overlap with the shared constructions.
As shown in Figure \ref{fig:exp_vs_name_vs_utterances}, we find that shared constructions are more similar to post- than to pre-interaction names. %
The plot also shows that the similarity of shared constructions with post-interaction names is higher for shared constructions used in the later rounds of the dialogue (Spearman’s $\rho= 0.2, p \ll 0.001$). Besides this recency effect, we also observe a frequency effect: the more frequently a shared construction is used by a participant, the more similar their post-interaction name will be to that construction (Spearman’s $\rho = 0.45, p \ll 0.001$).

Taken together, these results suggest that referential labels used individually after the interaction are strongly associated with shared constructions forged in interaction.

%% file: sections/5-analysis3.tex
\section{Analysis 3: Naming Convergence Across Speakers is Linked to Shared Constructions}

Finally, we investigate how similar the names individually given by each speaker to the fribbles before and after the interaction are to each other. \textcite{rasenberg2022primacy} and \textcite{eijk2022cabb} showed that the level of cross-speaker name similarity increases after the communicative task. 
Here we ask: To what extent is this converging trend related to the patterns of use of shared constructions?

To address this question, we first compute the cosine similarity between the two pre-interaction names ($S_{pre}$) 
and between the two post-interaction names ($S_{post}$) given by the participants and then measure the difference between these two similarities ($S_{post} - S_{pre}$). 
For actual participant pairs, we find that the mean cosine similarity between the two pre-interaction names is 0.06, while after the interaction the participant names have a mean cosine similarity of 0.43. Thus, there is an average increase of 0.37 cosine similarity. 
In contrast, for pseudo-pairs of participants, the average cosine similarity between the two post-interaction names is 0.07, and the similarity difference between pre- and post-interaction names is 0 in this case.
Hence, there is no increase of similarity in the pseudo-pairs. These results confirm that the converging trend in the post-interaction names already observed by \textcite{rasenberg2022primacy} and \textcite{eijk2022cabb} emanates from the interactive processes taking place during the communicative task, which are absent in the pseudo-pairs.

\begin{figure}[ht!]
    \centering
    \includegraphics[width=0.99\linewidth]{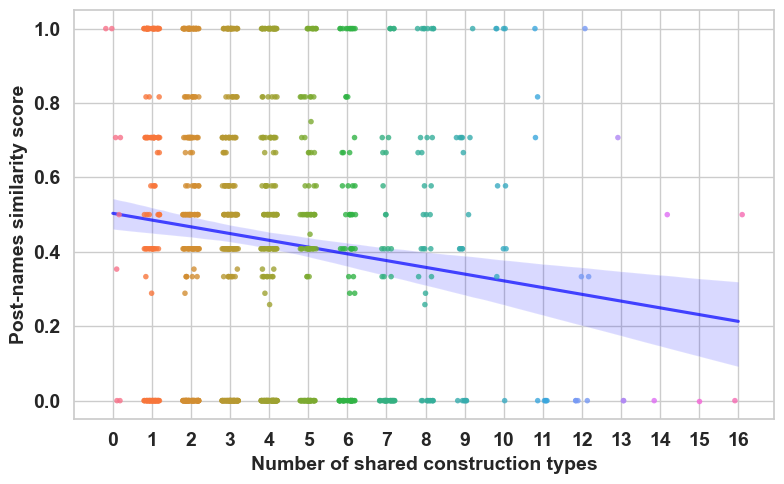}
    \caption{%
    Correlation between the number of shared construction types per object and the cosine similarity between the post-interaction names of the two participants in each dyad.}
    \label{fig:num_of_shared_exp_types_vs_similarity}
\end{figure}

Recall that our first analysis revealed that speakers may not immediately converge on a unique, simple %
shared construction type. 
Instead, they may use several complementary descriptions and entertain different alternative views of a referent, %
as illustrated by the example in Figure \ref{fig:shared_expression_types}. 
We hypothesise that using many different shared construction types may be indicative of difficulty building common ground and finding a simple way to refer to an object, which could lead to less similar post-interaction names.
Indeed, we find a weak negative correlation between the number of shared construction types and the cosine similarity of the post-interaction names (Spearman's $\rho=-0.13, p \ll 0.001$): the more construction types for a fribble, the less similar the post-interaction names tend to be, as shown in Figure \ref{fig:num_of_shared_exp_types_vs_similarity}.

Leaving aside the possible presence of various shared construction types per referent, we analyse two features of the most dominant shared construction type per fribble within a dialogue (e.g., ``boiler'' in Figure \ref{fig:shared_expression_types}): its frequency, i.e., in how many dialogue rounds it is used, and its recency, i.e., how close in terms of rounds the construction is to the end of the dialogue (and hence to the post-interaction naming task). We find that frequency positively correlates with the degree of cosine similarity between the participants' post-interaction names (Spearman’s $\rho = 0.28, p \ll 0.001$). That is, the more rounds the dominant shared construction is used in, the more similar the two post-interaction names are to each other. A weaker recency effect is also present: the later in the dialogue the dominant shared construction is used, the higher the degree of post-interaction similarity (Spearman’s $\rho = 0.17, p \ll 0.001$). 

In summary, this analysis shows that the degree of convergence between speakers %
post-interaction is mediated by the patterns of reuse captured by shared constructions. More construction types are associated with less convergence, while the frequency and recency of the most prominent type correlate with higher cross-speaker naming similarity.

%% file: sections/6-discussion.tex
\section{Discussion}
We study the reuse of linguistic behaviours in referential communication through the lens of automatically detected shared lemmatised constructions (or shared constructions for short), further grouped into shared construction types based on their common content lemmas. %
Our work examines the patterns of shared constructions and studies whether they mediate speakers' convergence in naming conventions for novel referents after social interaction.

\subsection{Dynamics of Shared Constructions and Cross-Speaker Convergence}
First, the study replicates the earlier findings that linguistic alignment is prevalent in referential communication, and we show this is not simply attributable to objects soliciting similar expressions as pseudo-pairs do not linguistically align in this way. 
The analysis also shows that speakers tend to entertain several shared construction types and eventually converge to fewer ones over the course of the conversation, thus discontinuing the use of aligned-upon content words. This finding goes beyond previous studies which have shown that speakers often simplify referential expressions by dropping hedges (e.g., ``like'' or ``sort of'') \autocite{clark1986referring, carroll1980naming} and closed-class parts of speech (e.g., pronouns, conjunctions and determiners)  \autocite{hawkins2020characterizing}.
Reducing the number of content-word-based shared constructions over the interaction might be linked to establishing a stronger common ground. The third analysis reveals that using a higher number of shared construction types could hinder the establishment of a single, straightforward naming convention. This is evident by the negative relationship between the number of shared construction types and speakers' post-interaction naming similarity. 
Nevertheless, when considering the most prominent shared construction type, higher usage frequency correlates with higher cross-speaker naming similarity. Thus, the degree of cross-speaker conventionalisation post-interaction is affected by a complex interplay between different properties of shared constructions.

\subsection{Individual Conventionalisation through Shared Constructions}
After the interaction, individual speakers tend to use labels that are strongly related to shared construction types that emerge during the interaction. 
Our findings indicate that shared construction types become increasingly similar to speakers' post-interaction names in later rounds of the interaction, showing a recency effect.
We also observe a strong frequency effect, with individual post-interaction names exhibiting higher similarity with more frequent shared constructions. Despite the observed low similarity between pre-interaction names and shared constructions, pre-interaction names may still be of importance for the conventionalisation process---a preliminary analysis not reported in this paper suggests that pre-interaction names are more similar to shared constructions types than to utterances that do not contain shared constructions. Future work could further investigate how pre-conceptions undergo joint transformation through interaction.

\subsection{Automatic Detection of Shared Constructions in Referential Communication}
We present a novel methodology that automatically captures linguistic alignment, complementing manual approaches used in previous studies such as those by \textcite{brennan1996conceptual} and \textcite{rasenberg2022primacy}, which are hard to scale to representative samples. For example, due to the time-consuming nature of manual coding, the study by \textcite{rasenberg2020alignment} limited their focus to the emergence of linguistic alignment in the first round of the dialogue.
In contrast, this study offers methods to facilitate the detection of the emergence \textit{and} dynamics
of linguistic alignment over time.

Although the methodology presented in this paper provides a valuable tool for studying the reuse of linguistic behaviours in referential communication, it comes with a few limitations. Firstly, it relies on manually transcribed conversations, which may be affected by spelling errors or transcript variations (which are limited in the CABB corpus by using strict transcription protocols and a minimal number of transcribers).
Second, one of the caveats in our study is that fribbles comprise a base plus 3 to 6 sub-parts. We observe that speakers commonly start out by describing various sub-parts, but over time they focus on the most salient ones. This might explain why they gradually drop shared construction types and converge to fewer ones as the interaction progresses. This observation also raises a question about whether speakers' shared constructions tend to become more holistic in nature by the end of the interaction (e.g., describing the whole of base plus sub-parts as ``boiler''). However, we lack information regarding the specific subpart(s) for which shared constructions are employed. Consequently, we cannot distinguish between cases where participants use a variety of holistic labels or multiple labels for different subparts. These are unanswered questions that can be explored in future studies.

Finally, an interesting future direction could be to explore whether the dynamics of the captured linguistic alignment during the conversation can account for changes in individual conceptualisations of fribbles (e.g., people might come to ``see'' a fribble as a robot as a result of the conversational history). To this end, behavioural and neural metrics of pre- to post-interaction change in conceptual representations of fribbles made available by \textcite{eijk2022cabb} could be exploited.

%% file: sections/7-conclusion.tex
\section{Conclusion}

This paper investigates how shared linguistic behaviours influence speakers' convergence on labels for novel objects after social interaction. We present an automated method to detect these behaviours, i.e., shared lemmatised constructions, and apply it to a referential communication corpus. The study shows a prevalence of shared constructions in the interactive communicative task. Particularly, speakers use a higher number of shared construction types in initial rounds, which decreases significantly as the interaction progresses.
As discussed above, our findings show a negative correlation between the number of shared construction types per referent and the similarity of speakers' post-interaction names. We thus find that more alignment is not always better when it comes to referring to novel objects; aligning on many referential expressions may, in fact, hinder speakers' convergence on an economic naming convention. Instead,  we find that using fewer shared construction types (and using those frequently and towards the end of the interaction) is more likely to result in convergence on an object label, as measured post-interaction.  The current %
approach to shared referential construction promises to help accelerate our understanding of how a common ground is carved out of joint action. 
This common ground building is ultimately related to how meaning is collaboratively negotiated in interaction from the bottom up. 
In related work \autocite{Akamine2024sp}, we make further progress in this direction by exploiting the present approach to analyse the interplay between lexical and gestural alignment.